# 3DYoga90: A Hierarchical Video Dataset for Yoga Pose Understanding

Seonok Kim
Korea University
145 Anam-ro, Seongbuk-gu, Seoul
`sokim0991@korea.ac.kr`

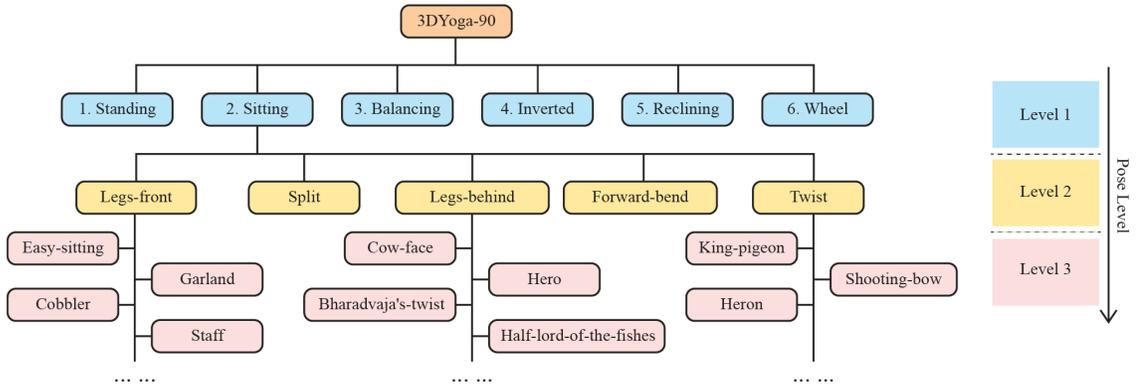

Figure 1. A Three-level Hierarchical Structure of 3DYoga90 Dataset. There are three levels of categorical labels: level 1, level 2, and level 3. In level 1, there are six classes; in level 2, twenty classes; and in level 3, ninety classes. The tree diagram displays example poses at level 2 and level 3 within the level 1 'Sitting' pose.

## Abstract

*The increasing popularity of exercises including yoga and Pilates has created a greater demand for professional exercise video datasets in the realm of artificial intelligence. In this study, we developed 3DYoga90[1], which is organized within a three-level label hierarchy. We have expanded the number of poses from an existing state-of-the-art dataset, increasing it from 82 to 90 poses. Our dataset includes meticulously curated RGB yoga pose videos and 3D skeleton sequences. This dataset was created by a dedicated team of six individuals, including yoga instructors. It stands out as one of the most comprehensive open datasets, featuring the largest collection of RGB videos and 3D skeleton sequences among publicly available resources. This contribution has the potential to significantly advance the field of yoga action recognition and pose assessment. Additionally, we conducted experiments to evaluate the practicality of our proposed dataset. We employed three different model variants for benchmarking purposes.*

## 1. Introduction

In recent times, the emergence of professional sports datasets [1-3, 5-14] has played a vital role in advancing research and applications in the field of health and well-being. Notably, the task of recognizing yoga poses presents a unique challenge, as it involves a diverse range of continuous movements, even within a single posture. Consequently, the demand for comprehensive datasets that capture these nuanced action units has become increasingly apparent. However, it is currently apparent that there is a shortage of open video datasets that offer fine-grained classes suitable for effective model training.

This deficiency prompted us to undertake the development of 3DYoga90. The root causes of the previous issue of data insufficiency can be attributed to two major factors: the requirement for expertise in a specific domain and the scarcity of data available on the web. We addressed these challenges by recruiting three yoga instructors to ensure the precision of pose categorization and data collection. To enhance data diversity, we conducted searches in various languages, including English, Korean, Japanese, Sanskrit, among others. Sanskrit is the classical Indian language used in yoga for its precision in conveying original texts and concepts. This enables us to collect data generated from different countries.

---

[1] Dataset and codes are available https://github.com/seonokkim/3DYoga90

| Dataset | Classes | Samples | Type | Source |
| --- | --- | --- | --- | --- |
| Yoga-82 [1] | 82 | 28,478 | RGB Image | Bing |
| 3D-Yoga [2] | 117 | 3,782 | RGB-D image, Skeleton | Mocap (Kinects) |
| YogaTube [3] | 82 | 5,484 | RGB video, Optical flow, Skeleton | YouTube, Bing, Flicker, etc. |
| **3DYoga90** | **90** | **6,177/5,526** | **RGB Video, 3D Skeleton** | **YouTube** |

Table 1. Comparison with the state-of-the-art yoga datasets. Recent yoga pose datasets are mostly comprised of data collected from the web. 3DYoga90 is the largest dataset, providing 6,177 RGB video samples and 5,526 3D skeleton samples.

The 3DYoga90 dataset employs a three-level label hierarchy, with levels 1 and 2 aligning with the Yoga-82 [1] dataset, while level 3 introduces an additional nine poses. Instead of being a 2D image dataset like Yoga-82 [1], our dataset consists of RGB yoga pose videos and 3D skeleton sequences extracted from YouTube. A dedicated team of six individuals, including yoga instructors, worked diligently to curate the dataset, ensuring the highest quality and accuracy. Figure 3 shows an overview of the 3DYoga90 dataset creation process.

This dataset that we have developed stands out as one of the most substantial open datasets, offering the largest collection of RGB videos and 3D skeleton sequences among publicly available resources.

Furthermore, to assess the practical utility of 3DYoga90, we conducted a series of experiments, employing benchmarking techniques with three distinct variants of a Deep Neural Network (DNN) model.

In summary, our contributions are:
- We introduce the largest dataset of yoga pose videos.
- Our video dataset comprises a comprehensive collection of 90 distinct poses organized into a three-level hierarchy.
- We've established a skeleton dataset, allowing for resource-efficient learning.

These contributions can promise to significantly advance the landscape of research and development in yoga action recognition and pose assessment.

## 2. Related Work

### 2.1. Yoga Datasets

Amid increasing interest in health and well-being, a variety of yoga-related datasets have recently emerged, catering to the growing demand for yoga-related research and applications.

Yoga-82 [1] is a hierarchical 2D dataset designed for large-scale yoga pose recognition with 82 classes. Each image contains one or more people performing the same yoga pose. The dataset comprises a total of 28,478 RGB images, sourced from Bing, a search engine.

3D-Yoga [2] consists of over 3,792 action samples and 16,668 RGB-D key frames from 22 subjects performing 117 yoga poses with two RGB-D cameras. It includes reconstructed 3D yoga poses, and experiments with the Cascade 2S-AGCN show superior accuracy in pose recognition and assessment.

YogaTube [3] introduces a new large-scale video benchmark dataset for yoga action recognition, featuring 82 classes and 5,484 samples. It includes RGB videos, optical flow data, and skeleton data from sources such as YouTube, Bing, Flickr, and more.

Our 3DYoga90 dataset comprises 90 classes and a total of 6,177 samples, encompassing RGB videos and 3D skeleton data sourced from YouTube. Distinguished as one of the most extensive open datasets available, it boasts the largest number of samples among comparable datasets. This significant contribution aims to enhance the research and development of yoga action recognition and pose assessment. Table 1 shows the detailed comparison.

### 2.2. Web-based Video Dataset

To offer a video dataset for users to download, one must possess substantial storage capacity. However, recent publications in top-tier conferences related to Video Datasets frequently highlight the advantages of harnessing. the abundance of publicly available videos on the internet. The authors of FineGym [5], WLASL [15] and MS-ASL [35] do not store raw video datasets on online repositories for download; instead, they provide JSON files containing web-based resources such as YouTube links or video identifiers. They also offer code for downloading the respective videos, allowing providers and users to access and store only the necessary data. Furthermore, by specifying the starting and ending points of specific segments related to classes, they enable the extraction of fine-grained video clips. Inspired by these approaches, we have also created our dataset by utilizing existing web-based videos like YouTube.

| Type | Standing | Sitting | Balancing | Inverted | Reclining | Wheel |
|------|----------|---------|-----------|----------|-----------|-------|
| RGB videos | 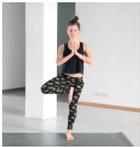 | 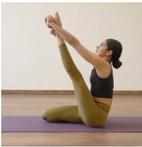 | 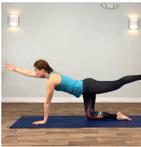 | 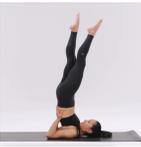 | 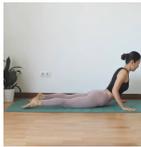 | 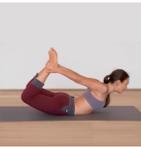 |
| 3D sequences | 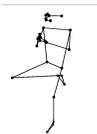 | 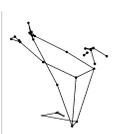 | 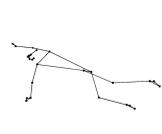 | 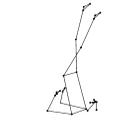 | 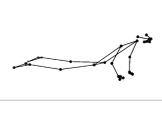 | 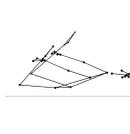 |

Table 2. Examples from the 3DYoga90 Dataset. The dataset comprises two types of data: RGB videos and 3D skeleton sequences. The number of RGB videos is 6,177, and the number of 3D skeleton sequences is 5,482. From left to right, the examples are tree, heron, balancing table, shoulder stand, cobra, and bow poses.

| Level 1 | Level 2 | Level 3 |
|---------|---------|---------|
| 1. Standing | 11-14 (4 Classes) | 101-125 (25 Classes) |
| 2. Sitting | 15-19 (6 Classes) | 126-143 (18 Classes) |
| 3. Balancing | 20-21 (2 Classes) | 144-152 (9 Classes) |
| 4. Inverted | 22-23 (2 Classes) | 153-158 (18 Classes) |
| 5. Reclining | 24-27 (4 Classes) | 159-179 (63 Classes) |
| 6. Wheel | 28-30 (3 Classes) | 180-190 (33 Classes) |
| Total | 20 Classes | 90 Classes |

Table 3. An overview of 3DYoga90 Dataset. Level 1 consists of 6 classes, level 2 has 20 classes, and level 3 has 90 classes. Reclining has the most poses with 63, and Balancing has the fewest with 9.

### 2.3. RGB and Skeleton-based Action Recognition

Human Action Recognition (HAR) involves the collection and processing of human action information in various ways to recognize and classify actions. There are various types of data for performing HAR, broadly categorized into image-based data and non-image data. Image-based data includes RGB, 3D skeletons, depth, infrared, point cloud, event streams, and more. In this context, we provide data specifically for the mainstream tasks of RGB and 3D skeleton-based action recognition.

For RGB-based action recognition, CNN-based models [22, 23, 24] are commonly used. Among these, 3D CNN [36] is a model that extends 2D CNNs to learn in the temporal dimension. It operates on input data structured as temporal, channel, width, and height dimensions, followed by action recognition.

In the case of 3D skeleton-based action recognition, the widely used models are models based on DNN [37, 38, 40], RNN [29-34], and GCN [25-28]. ST-GCN model [28] is one of the most frequently used models. It stacks skeleton data over time and performs action recognition through graph operations. To perform 3D skeleton-based action recognition, skeletons composed of keypoints are detected through pose estimation from RGB-based videos. This information is then fed into the model to predict the final action class. We validated the feasibility of our introduced dataset using a basic DNN model.

## 3. 3DYoga90 Dataset

### 3.1. RGB Video Collection

Six individuals, including three yoga instructors and three non-instructors, collaborated on this effort to curate a dataset of RGB yoga pose videos from YouTube.

Between June 2023, and July 2023, we conducted searches in multiple languages, including English, Korean, and Sanskrit, with the primary objective of identifying videos that showcased specific yoga poses (classes). For each video, we meticulously recorded the start and end frames within a 30-second time window. Our primary data criterion was to ensure the visibility of all 33 skeleton points. We deliberately selected videos that provided a comprehensive view of the entire body and featured uninterrupted sequences, including a side view of the pose whenever possible. In the following cases, we excluded data from our collection: (1) when poses did not appear continuously within a single video, such as when close-up shots were included in the middle, (2) when there were two or more individuals appearing in the video, and (3) when significant landmarks were situated outside the frame or obscured by visual effects. We collected raw videos for lengths of 30 minutes or less. Subsequently, we extracted the segments containing the targeted pose from each raw video by specifying start and end frames. The duration between the start and end frames was constrained to be within 30 seconds. Figure 2 illustrates the distribution of the duration for each pose, adhering to this criterion.

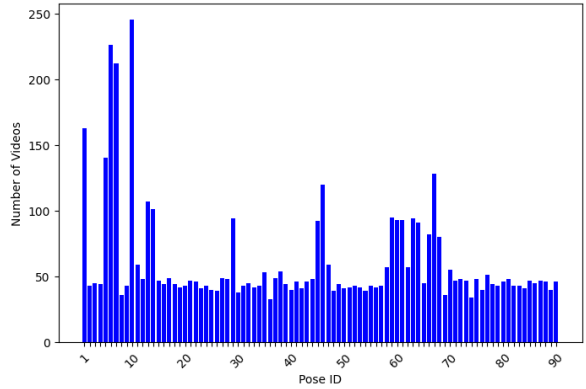

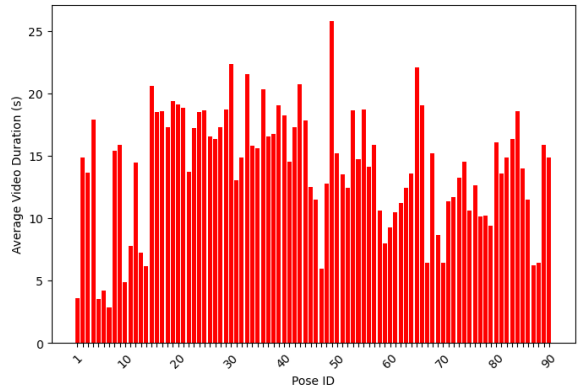

Figure 2. Dataset statistics of 3DYoga90 dataset. Pose ID refers to the pose identification at level 3. Each pose has an average of 75 videos, and the average sequence duration is 12 seconds.

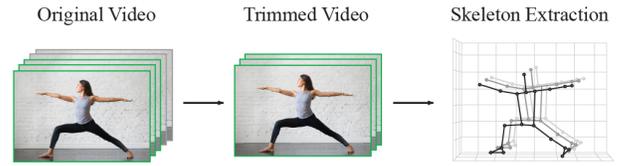

Figure 3. An overview of the 3DYoga90 dataset creation process. After extracting relevant segments from raw videos collected on the web, landmark information is utilized to generate skeleton data.

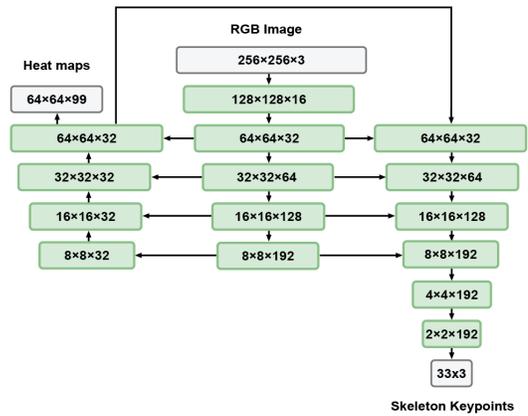

Figure 4. Network architecture of BlazePose [4]. The 3DYoga90 skeleton dataset consists of (x, y, z) keypoints for 33 landmarks in trimmed videos, and these keypoints are predicted using BlazePose.

### 3.2. Skeleton Dataset

Our skeleton dataset generation process consists of two main phases: data extraction, and transformation.

**Data Extraction** The first phase involves extracting relevant pose data from video files. The core of our data extraction process is the integration of BlazePose [4] model. This model efficiently detects and tracks human body landmarks, providing us with accurate 3D coordinates (x, y, z) for each landmark in every frame of the video.

BlazePose [4] is a lightweight network for real-time human pose estimation on mobile devices. The model predicts the locations of all 33 key points, using a combined heatmap, offset, and regression approach. During training, they employ heatmap and offset loss, removing the corresponding output layers for inference. This approach effectively utilizes heatmaps to supervise lightweight embedding, enhancing coordinate regression accuracy. Additionally, skip-connections are used to balance high and low-level features within the network, further improving the model's performance.

Through this model, we extract 33 key points data for each frame in every video, across $x$, $y$, and $z$. This phase ensures that we capture a comprehensive representation of human poses.

**Data Transformation** Once the pose data is extracted, we move on to the data transformation phase. Here, we systematically organize the extracted data into a structured format, creating a well-defined data frame. This tabular representation includes columns such as the number of frames, and coordinates. By structuring the data in this manner, we facilitate efficient management and analysis of the pose information. Researchers can easily access and manipulate the dataset for their specific needs, making it a valuable resource for the field of human motion tracking and analysis.

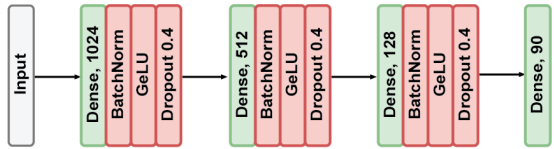

Figure 5. An instantiation of our 3D poses estimation architecture for DNN variant 3. We configured the initial dense layer with 1024 output channels, followed by batch normalization, activation, and dropout with a dropout rate set to 0.4.

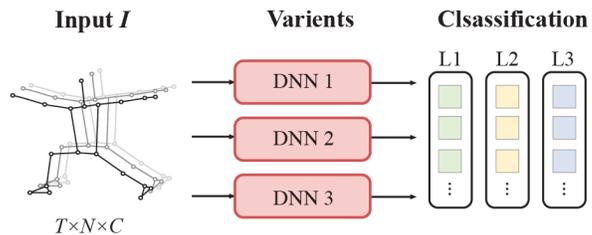

Figure 6. The pipeline of our proposed yoga pose classification. An input skeleton sequence $I$ was fed into three different variants of a Deep Neural Network (DNN) to predict level 1, 2, and 3 poses and the results were compared. $T$ represents the number of frames in a sequence, $N$ denotes the number of keypoints, and $C$ indicates 3 channels representing $(x, y, z)$.

Following these phases, we save the processed data in Parquet format. During the creation of the skeleton dataset, the variant of the BlazePose [4] model may miss capturing poses in some video frames. As a result, while the RGB video dataset comprises 6,177 videos, the skeleton dataset is filtered down to 5,526 sequences.

### 3.3. Label Hierarchy and Annotation

In constructing our dataset, we primarily adopt the hierarchy proposed in Yoga-82 [1], which serves as our foundational framework. However, we have extended this framework by incorporating an additional eight poses suggested by YogaRenew [39], an Online Yoga Teacher Training institution, resulting in a comprehensive dataset.

Our dataset's structure involves three distinct levels. At level 1, we have six primary categories, including Standing, Sitting, Balancing, Inverted, Reclining, and Wheel. These level 1 classes provide a broad classification of yoga poses.

Level 2 introduces subcategories within each level 1 category. For example, 'Standing' includes 11 to 14 classes, 'Sitting' encompasses 15 to 19 classes, 'Balancing' consists of 20 to 21 classes, 'Inverted' covers 22 to 23 classes, 'Reclining' involves 24 to 27 classes, and 'Wheel' comprises 28 to 30 classes. This subcategorization enhances the granularity of our classification.

Level 3 represents fine-grained classes. Within each Level 2 subcategory, we provide further refinement. For example, within the 'Standing' subcategory, we have a diverse set of classes ranging from 101 to 125, offering detailed classification.

### 3.4. Dataset Statistics

We extracted a total of 5,482 skeleton sequences by trimming clips from 6,177 original videos collected from 2,170 different sources spanning various countries. The distribution of level 3 poses is presented in Figure 2, where the 'Downward Dog' pose emerges as the most common, with 245 videos. On average, each pose is represented in 75 videos.

The overall mean sequence duration is approximately 12.16 seconds. Within the level 3 pose class, sequence durations vary from a minimum of 2.83 seconds for the 'Halfway Lift' pose to a maximum of 25.84 seconds for the 'Balancing Table' pose. Detailed information on the distribution of level 3 pose class durations can be found in Figure 2.

## 4. Experiments

### 4.1. Benchmarking 3DYoga-90 Dataset

To assess the utility of our proposed dataset, we conducted training and evaluation following the class hierarchy of the dataset. We focused on the Skeleton dataset, as limitations in computation resources prevented us from experimenting with RGB videos. We applied a Deep Neural Network (DNN) model to predict the three levels of our proposed pose dataset.

The DNN model consists of three variants, namely DNN-Small, DNN-Base and DNN-Large. DNN-Large model is described in Figure 5.

**DNN-Small:** DNN-Small model includes an Input Layer, followed by a Dense layer that yields 1024 outputs.

| Level | DNN | Prec. | Recall | F1 | Acc. |
|---|---|---|---|---|---|
| L1 | Small | 0.945 | 0.942 | 0.943 | 0.9453 |
| L1 | Base | **0.947** | **0.946** | **0.946** | **0.9507** |
| L1 | Large | 0.945 | 0.940 | 0.942 | 0.9489 |
| L2 | Small | 0.880 | 0.875 | 0.873 | 0.9088 |
| L2 | Base | 0.894 | 0.876 | 0.880 | 0.9106 |
| L2 | Large | 0.874 | 0.863 | 0.864 | 0.9051 |
| L3 | Small | 0.850 | 0.843 | 0.832 | 0.8631 |
| L3 | Base | 0.856 | 0.851 | 0.837 | 0.8668 |
| L3 | Large | 0.842 | 0.833 | 0.818 | 0.8631 |

Table 4. Action recognition results of three DNN model variants. Overall, the prediction results for the level 1 pose are better. The DNN-Base model has the best pose prediction performance for level 1.

This is followed by a Batch Normalization layer, an Activation layer, and a Dropout layer. Subsequently, there is another Dense layer that generates 512 outputs, followed by Batch Normalization, Activation, and Dropout layers. Finally, the model concludes with a Dense layer that produces 90 outputs.

**DNN-Base:** DNN-Base is a modified version of DNN-Small, where an additional layer with 256 outputs has been added to the end of the architecture.

**DNN-Large:** DNN-Large is an extension of DNN-Base, with the addition of a final section consisting of a Dense layer with 128 outputs, followed by Batch Normalization, Activation, and Dropout layers.

To evaluate the effectiveness of our three-level hierarchical classification task, we meticulously designed experiments. For this purpose, we carefully curated label sets, consisting of 6, 20, and 90 classes that corresponded to levels 1, 2, and 3, respectively.

Within our data preprocessing pipeline, we introduced a custom layer to handle various feature operations, such as averaging, concatenation, padding for segmentation, flattening for dimension reduction, resizing, and data aggregation. These operations were pivotal in generating comprehensive feature vectors from the input data, ensuring uniform dimensions, and preserving crucial statistical information. These feature vectors served as the foundation for subsequent model training and analysis.

During model training, we utilized a batch size of 64 to optimize gradient updates effectively. To mitigate overfitting and evaluate model performance, we incorporated a 10% validation split. The learning rate was set at 3.33e-4, with a learning rate reduction factor of 0.8 for fine-tuning model weights. Our default training configuration consisted of 100 epochs, with an early stopping mechanism in place to halt training as needed.

## 4.2. Results

In analysing the experimental results, several key observations emerge. At level 1, all three variants, DNN-Small, DNN-Base, and DNN-Large, exhibit strong performance, boasting high precision, recall, F1 scores, and accuracy. DNN-Base particularly stands out with the highest accuracy, indicating its proficiency in level 1 classification.

Transitioning to level 2, the overall performance remains robust across all three variants, albeit with a slight decrease in precision and recall compared to Level 1. Nonetheless, DNN-Base continues to excel, maintaining the highest accuracy and F1 score.

However, level 3 presents a notable challenge due to its requirement to classify a more extensive set of 90 finely grained poses. As anticipated, this increased granularity presents a more intricate problem compared to level 1 and 2. DNN-Base maintains its superiority in terms of accuracy even in this more demanding context.

In summary, all three variants consistently perform well, with DNN-Base consistently outperforming others in both level 1 and level 2 classification tasks. The drop in performance metrics at level 3 underscores the increased complexity of classifying a larger number of fine-grained poses.

## 5. Conclusion

In this paper, we introduce the 3DYoga90 dataset. Featuring a three-level label hierarchy and a substantial repository of RGB videos and 3D skeleton sequences, this dataset offers considerable promise for the advancement of yoga action recognition and pose assessment. Our experiments, involving three distinct DNN model variants, consistently demonstrated strong performance, with DNN-Base excelling particularly in level 1 and level 2 tasks. Consequently, the 3DYoga90 dataset is well-positioned to leave a substantial mark in this research domain.

Further research is warranted to address the heightened complexity of level 3 poses. Our future studies should encompass not only the skeleton classification task but also assess the performance in RGB video action recognition tasks.

In the future, we will conduct evaluations of performance using an RGB-based dataset since we conducted benchmark experiments using a skeleton-based dataset. Furthermore, we plan to assess the dataset's validity by incorporating state-of-the-art graph-based models.

## Acknowledgements

We are deeply grateful to the data collection and annotation team including professional yoga instructors. Special thanks to Jaehun Yang. This research is a collaborative effort, and we thank all who contributed.

# References

[1] Manisha Verma, Sudhakar Kumawat, Yuta Nakashima, and Shanmuganathan Raman. Yoga-82: A New Dataset for Fine-grained Classification of Human Poses. In *Proceedings of IEEE/CVF Conference on Computer Vision and Pattern Recognition Workshops (CVPRW)*, 2020, pp. 4472-4479.

[2] Jianwei Li, Haiqing Hu, Jinyang Li, Xiaomei Zhao. 3D-Yoga: A 3D Yoga Dataset for Visual-based Hierarchical Sports Action Analysis. In *Proceedings of the Asian Conference on Computer Vision (ACCV)*, 2022, pp. 434-450.

[3] Santosh Kumar Yadav, Guntaas Singh, Manisha Verma, Kamlesh Tiwari, Hari Mohan Pandey, Shaik Ali Akbar, and Peter Corcoran. YogaTube: A Video Benchmark for Yoga Action Recognition. In *Proceedings of the International Joint Conference on Neural Networks (IJCNN)*, 2022.

[4] Valentin Bazarevsky, Ivan Grishchenko, Karthik Raveendran, Tyler Zhu, Fan Zhang, and Matthias Grundmann. BlazePose: On-device Real-time Body Pose Tracking. *arXiv preprint arXiv:2006.10204*, 2020.

[5] Dian Shao, Yue Zhao, Bo Dai, and Dahua Lin. FineGym: A Hierarchical Video Dataset for Fine-grained Action Understanding. In *Proceedings of the IEEE/CVF Conference on Computer Vision and Pattern Recognition (CVPR)*, 2020.

[6] Dittakavi Bhat, Divyagna Bavikadi, Sai Vikas Desai, Soumi Chakraborty, Nishant Reddy, Vineeth N Balasubramanian, and Bharathi Callepalli. Pose Tutor: An Explainable System for Pose Correction in the Wild. In *Proceedings of the IEEE/CVF Conference on Computer Vision and Pattern Recognition Workshops (CVPRW)*, pages 3539-3548, 2022.

[7] Khurram Soomro, Amir Roshan Zamir, and Mubarak Shah. UCF101: A Dataset of 101 Human Actions Classes from Videos in The Wild. *arXiv preprint arXiv:1212.0402*, 2012.

[8] Yingwei Li, Yi Li, Nuno Vasconcelos. RESOUND: Towards Action Recognition without Representation Bias. In *Proceedings of the European Conference on Computer Vision (ECCV)*, 2018, pp. 513-528.

[9] Andrej Karpathy, George Toderici, Sanketh Shetty, Thomas Leung, Rahul Sukthankar, Li Fei-Fei. Large-scale Video Classification with Convolutional Neural Networks. In *Proceedings of the IEEE Conference on Computer Vision and Pattern Recognition (CVPR)*, 2014.

[10] William McNally, Kanav Vats, Tyler Pinto, Chris Dulhanty, John McPhee, and Alexander Wong. GolfDB: A Video Database for Golf Swing Sequencing. In *Proceedings of the IEEE/CVF Conference on Computer Vision and Pattern Recognition Workshops (CVPRW)*, 2019, pp. 0-0.

[11] Khurram Soomro and Amir Roshan Zamir. Action Recognition in Realistic Sports Videos. In *Computer vision in sports*, 2013, pp. 181-208.

[12] Hamed Pirsiavash, Carl Vondrick and Antonio Torralba. Assessing the quality of actions. In *Proceedings of the European Conference on Computer Vision (ECCV)*, 2014, pp. 556–571.

[13] Paritosh Parmar and Brendan Tran Morris. What and How Well You Performed? A Multitask Learning Approach to Action Quality Assessment. In *Proceedings of the IEEE/CVF Conference on Computer Vision and Pattern Recognition (CVPR)*, 2019, pp. 304–313

[14] Shenlan Liu, Xiang Liu, Gao Huang, Lin Feng, Lianyu Hu, Dong Jiang, Aibin Zhang, Yang Liu and Hong Qiao. FSD-10: A Dataset for Competitive Sports Content Analysis. In *Neurocomputing,* 2020, pp. 360–367

[15] Dongxu Li, Xin Yu, Chenchen Xu, Lars Petersson and Hongdong Li. Transferring Cross-domain Knowledge for Video Sign Language Recognition. In *Proceedings of the IEEE/CVF Conference on Computer Vision and Pattern Recognition (CVPR)*, 2020.

[16] Guilhem Cheron, Ivan Laptev and Cordelia Schmid. P-CNN: Pose-Based CNN Features for Action Recognition. In *Proceedings of the IEEE/CVF International Conference on Computer Vision (ICCV)*, 2015.

[17] Qiuhong Ke, Mohammed Bennamoun, Senjian An, Ferdous Sohel and Farid Boussaid. A New Representation of Skeleton Sequences for 3D Action Recognition. In *Proceedings of the IEEE/CVF Conference on Computer Vision and Pattern Recognition (CVPR)*, 2017, pp. 3288–3297.

[18] Bo Li, Mingyi He, Xuelian Cheng, Yucheng Chen and Yuchao Dai. In *Proceedings of the IEEE International Conference on Multimedia and Expo Workshops (ICMEW)*, 2017, pp. 601–604.

[19] Hong Liu, Juanhui Tu and Mengyuan Liu. Two-Stream 3D Convolutional Neural Network for Skeleton-Based Action Recognition. *arXiv preprint arXiv:1705.08106*, 2017.

[20] Du Tran, Lubomir Bourdev, Rob Fergus, Lorenzo Torresani and Manohar Paluri. Learning Spatiotemporal Features with 3D Convolutional Networks. In *Proceedings of the IEEE/CVF International Conference on Computer Vision (ICCV)*, 2015, pp. 4489–4497.

[21] Haodong Duan, Yue Zhao, Kai Chen, Dahua Lin, Bo Dai. Revisiting Skeleton-based Action Recognition. In *Proceedings of the IEEE/CVF Conference on Computer Vision and Pattern Recognition (CVPR)*, 2022.

[22] Haodong Duan, Yue Zhao, Kai Chen, Dahua Lin, Bo Dai. Revisiting Skeleton-based Action Recognition. In *Proceedings of the IEEE/CVF Conference on Computer Vision and Pattern Recognition (CVPR)*, 2022, pp. 2969-2978.

[23] Christoph Feichtenhofer. X3D: Expanding Architectures for Efficient Video Recognition. In *Proceedings of the IEEE/CVF Conference on Computer Vision and Pattern Recognition (CVPR)*, 2020, pp. 203–213.

[24] Christoph Feichtenhofer, Haoqi Fan, Jitendra Malik, and Kaiming He. Slowfast networks for video recognition. In *Proceedings of the IEEE/CVF International Conference on Computer Vision (ICCV)*, 2019, pp. 6202–6211.

[25] Chen, Yuxin, Ziqi Zhang, Chunfeng Yuan, Bing Li, Ying Deng, and Weiming Hu. Channel-wise Topology Refinement Graph Convolution for Skeleton-Based Action Recognition. In *Proceedings of the IEEE/CVF International Conference on Computer Vision (ICCV)*, 2021, pp. 13359-13368.

[26] Ke Cheng, Yifan Zhang, Xiangyu He, Weihan Chen, Jian Cheng, and Hanqing Lu. Skeleton-Based Action Recognition With Shift Graph Convolutional Network. In *Proceedings of the IEEE/CVF Conference on Computer Vision and Pattern Recognition (CVPR)*, 2020, pp. 183-192.

[27] Ke Cheng, Yifan Zhang, Xiangyu He, Weihan Chen, Jian Cheng, and Hanqing Lu. Skeleton-Based Action

Recognition With Shift Graph Convolutional Network. In *Proceedings of the IEEE/CVF Conference on Computer Vision and Pattern Recognition (CVPR)*, 2019.

[28] Sijie Yan, Yuanjun Xiong, and Dahua Lin. Spatial temporal graph convolutional networks for skeleton-based action recognition. In *Proceedings of the AAAI conference on artificial Intelligence*, 2018.

[29] Jeff Donahue, Lisa Anne Hendricks, Sergio Guadarrama, Marcus Rohrbach, Subhashini Venugopalan and Trevor Darrell, Kate Saenko. In *Proceedings of the IEEE/CVF Conference on Computer Vision and Pattern Recognition (CVPR)*, 2015, pp. 2625–2634.

[30] Yong Du, Wei Wang and Liang Wang. Hierarchical recurrent neural network for skeleton based action recognition. In *Proceedings of the IEEE/CVF Conference on Computer Vision and Pattern Recognition (CVPR)*, 2015, pp. 1110–1118.

[31] Jun Liu, Amir Shahroudy, Mauricio Perez, Gang Wang, Ling-Yu Duan and Alex C. Kot. NTU RGB+D 120: A Large-Scale Benchmark for 3D Human Activity Understanding. In *Proceedings of the IEEE Transactions on Pattern Analysis and Machine Intelligence (TPAMI)*, 2020, pp. 2684-2701.

[32] Jun Liu, Amir Shahroudy, Dong Xu and Gang Wang. Spatio-Temporal LSTM with Trust Gates for 3D Human Action Recognition. In *Proceedings of the European Conference on Computer Vision (ECCV)*, 2016, pp. 816–833.

[33] Sijie Song, Cuiling Lan, Junliang Xing, Wenjun Zeng and Jiaying Liu. An End-to-End Spatio-Temporal Attention Model for Human Action Recognition from Skeleton Data. In *Proceedings of the AAAI Conference on Artificial Intelligence*, 2017, pp. 4263–4270.

[34] Pengfei Zhang, Cuiling Lan, Junliang Xing, Wenjun Zeng, Jianru Xue and Nanning Zheng. View Adaptive Neural Networks for High Performance Skeleton-based Human Action Recognition. In *Proceedings of the IEEE Transactions on Pattern Analysis and Machine Intelligence (TPAMI)*, 2019, pp. 1963–1978.

[35] Hamid Vaezi Joze and Oscar Koller, MS-ASL: A Large-Scale Data Set and Benchmark for Understanding American Sign Language, In *Proceedings of the British Machine Vision Conference (BMVC)*, 2019.

[36] Shuiwang Ji, W. Xu, Ming Yang and Kai Yu. 3D convolutional neural networks for human action recognition. In *Proceedings of the IEEE Transactions on Pattern Analysis and Machine Intelligence (TPAMI)*, 2012, pp. 221-231.

[37] Hung-Cuong Nguyen, Thi-Hao Nguyen, Rafał Scherer and Van-Hung Le. Deep Learning for Human Activity Recognition on 3D Human Skeleton: Survey and Comparative Study, *arXiv preprint arXiv:2305.15692*, 2023.

[38] Manh-Hung Ha and Oscal Tzyh-Chiang Chen. Deep Neural Networks Using Residual Fast-Slow Refined Highway and Global Atomic Spatial Attention for Action Recognition and Detection. In *IEEE Access*, 2021.

[39] YogaRenew. https://www.yogarenewteachertraining.com/online-yoga-teacher-training-courses/200-hour-online-yoga-teacher-training. Assessed: 2023-10-16.

[40] Robert Hatch. Kaggle Notebook. https://www.kaggle.com/code/roberthatch/gislr-lb-0-63-on-the-shoulders. Assessed: 2023-10-16.